\documentclass{article}

\usepackage{arxiv}

\usepackage[utf8]{inputenc} % allow utf-8 input
\usepackage[T1]{fontenc}    % use 8-bit T1 fonts
\usepackage{hyperref}       % hyperlinks
\usepackage{url}            % simple URL typesetting
\usepackage{booktabs}       % professional-quality tables
\usepackage{amsfonts}       % blackboard math symbols
\usepackage{nicefrac}       % compact symbols for 1/2, etc.
\usepackage{microtype}      % microtypography
\usepackage{graphicx}
\usepackage{doi}
\usepackage{amsmath}
\usepackage{subcaption}
\usepackage{tabularx}
\usepackage{float}

\title{Exploration and Evaluation of Bias in Cyberbullying Detection with Machine Learning}

\author{Andrew Root \\
	Department of Computer Science\\
	University of Wisconsin Eau Claire\\
	Eau Claire, WI 54702 \\
	\texttt{roota5351@uwec.edu} \\
	\And
	Liam Jakubowski \\
	Department of Computer Science\\
	University of Wisconsin Eau Claire\\
	Eau Claire, WI 54702 \\
	\texttt{jakubola6675@uwec.edu} \\
        \And
        Dr. Mounika Vanamala \\
        Department of Computer Science \\
        University of Wisconsin Eau Claire \\
        Eau Claire, WI 54702 \\
        \texttt{vanamalm@uwec.edu} \\
}

\renewcommand{\shorttitle}{Cyberbullying Detection with Machine Learning}

\hypersetup{
pdftitle={Exploration and Evaluation of Bias in Cyberbullying Detection with Machine Learning},
pdfsubject={cs},
pdfauthor={Andrew Root, Liam Jakubowski, Mounika Vanamala},
pdfkeywords={cyberbullying detection; machine learning; cross-dataset evaluation; biased data collection, dataset expansion methods},
}

\begin{document}
\maketitle

\begin{abstract}
\indent It is well known that the usefulness of a machine learning model is due to its ability to generalize to unseen data. This study uses three popular cyberbullying datasets to explore the effects of data, how it's collected, and how it's labeled, on the resulting machine learning models. The bias introduced from differing definitions of cyberbullying and from data collection is discussed in detail. An emphasis is made on the impact of dataset expansion methods, which utilize current data points to fetch and label new ones. Furthermore, explicit testing is performed to evaluate the ability of a model to generalize to unseen datasets through cross-dataset evaluation. As hypothesized, the models have a significant drop in the Macro F1 Score, with an average drop of 0.222. As such, this study effectively highlights the importance of dataset curation and cross-dataset testing for creating models with real-world applicability. The experiments and other code can be found at https://github.com/rootdrew27/cyberbullying-ml.
\end{abstract}

% keywords can be removed

\section{Introduction}
Detecting cyberbullying with machine learning (ML) involves dealing with many subtle facets. These facets include but are not limited to, differing definitions of cyberbullying \cite{Survey}, varying labeling schemes, and biased data collection \cite{davidson}\cite{Dataset1}, all of which introduce bias in the ML models created. Many attempts have been made to address these different forms of bias \cite{ID-XCB}\cite{AccurateCyberbullyingDetectionandPreventiononSocialMedia}\cite{swear_words}, but removing bias altogether remains difficult, if achievable. Regardless, measuring and understanding bias remains important for understanding when an ML model is useful. This study takes as its primary concern, the bias introduced from cyberbullying datasets and the applicability of ML for cyberbullying detection. The experiments done in this study are referred to as cross-dataset evaluations, which evaluate ML models on datasets unseen during the training process. These tests are beneficial when there is potential for introducing bias in the data collection and labeling steps of ML. This paper reiterates and explicitly shows that many cyberbullying datasets are only useful so long as the models trained with them make predictions on data points from that same dataset: this is a major concern. When a dataset isn’t useful for creating ML models that perform well on other datasets, this indicates a fundamental difference in the understanding of the task at hand, bias in the dataset, or an error in the methodology. 

\indent The identification of cyberbullying is a subjective task, as it is dependent upon the labeler and the definition of cyberbullying that the labeler uses. Definitions such as “willful and repeated harm inflicted through the medium of electronic text” used by Reynolds et al. \cite{Reynolds} and “an aggressive, intentional act carried out by a group or individual, using electronic forms of contact, repeatedly and over time, against a victim who cannot easily defend him or herself” used by Van Hee et al. \cite{game}, have notable differences. More definitions observed in the literature are documented in \cite{Survey}. Unsurprisingly, these differences impact the labeling process and the models themselves. With regard to the labeling scheme, many datasets are incompatible due to the nature of the labeling process. For example, one study assigned labels to Twitter users rather than individual tweets \cite{mean_birds}. This level of variation makes it difficult to utilize multiple datasets without major refactoring or relabeling. Often though, the variation in labeling schemes can be as simple as referring to cyberbullying as harassment \cite{Dataset3}. A less explicit form of bias is the bias from data collection methods. Data collection methods used to curate cyberbullying datasets frequently rely on a lexicon of keywords (e.g. the Hatebase.org lexicon) to obtain tweets that include offensive words \cite{Dataset3}\cite{davidson}\cite{AccurateCyberbullyingDetectionandPreventiononSocialMedia} \cite{hatespeech_in_social_media}, which are believed to be associated with cyberbullying. This leads to an over-emphasis on the keywords and disregards nuanced, less explicit forms of cyberbullying. For example, one study used knowledge of tweets they had currently labeled to pick 17 keywords relevant to cyberbullying and queried for more tweets containing these words \cite{waseem-hovy-2016-hateful}. This clearly brings additional bias into the dataset and unfortunately, this dataset has continued to see usage without acknowledgment of the bias \cite{Dataset1}\cite{ignoring_bias}. Furthermore, bias in data can also come from semi-supervised learning, or automated dataset expansion methods, as they not only collect data based on keywords but potentially apply a label to the new data points \cite{Dataset1}. This approach is formulated and discussed in greater detail in the Materials and Methods section.\\
\indent A significant challenge in this field is the poor cross-dataset performance, where performance declines substantially when models are applied to datasets different from those on which they were trained. This issue stems primarily from the lexicon-based data collection methods frequently used in cyberbullying research. Lexicons, or predefined lists of offensive terms, are commonly used to capture explicit abusive language. However, this reliance tends to overfit models to the specific language patterns of a single dataset, making them less effective on datasets where abusive language may be more nuanced or context-dependent. For example, Pitsilis et al. \cite{RNN} used recurrent neural networks combined with word frequency vectors and user-based features, achieving strong results within their original dataset but facing limitations when attempting to detect nuanced hate speech in others. Similarly, Ruwandika and Weerasinghe \cite{ID-XCB} observed that lexicon-dependent models performed well in binary hate/non-hate classifications in local datasets but struggled with non-explicit forms of cyberbullying in different contexts. These examples demonstrate that models highly attuned to one dataset’s language profile often fail to generalize to other datasets, highlighting the need for adaptable data collection strategies beyond lexicons to improve cross-dataset applicability.

\indent Improving the real-world applicability of machine learning models is the ultimate goal of this study and preventing, measuring, and interpreting bias throughout the machine learning process is an essential component of this goal. After analyzing many studies and experiments performed, we deemed it necessary to explore the causes and impacts of bias in machine learning, particularly with respect to bias in the data, as we believe this contributes to the goal of creating real-world applicable models.  
\newline

\section{Materials and Methods}
\label{sec:Materials_and_Methods}

\subsection{Datasets}
There are many important aspects of datasets, but this study focuses on the data collection process, the labeling process, and any augmentations that are applied. This section will briefly cover each of these aspects, for each dataset, and direct readers to additional information about the datasets. Furthermore, properties of the dataset relevant to the experiments performed in this study are highlighted. Aside from their differences, the datasets are all composed of Tweets from the social media platform Twitter, and the curators of each dataset were all attempting to detect, what can broadly be described as, cyberbullying.

\subsubsection{Dataset 1}
This dataset is an aggregate of seven cyberbullying datasets, with additional data points curated via an algorithmic process \cite{Dataset1}. The creation dates of the datasets span from 2012 to 2019. The aggregate dataset has six classes: Age, Ethnicity, Gender, Religion, Other Cyberbullying, and Not Cyberbullying. To make these classifications the researchers collected positive and negative classes of each original dataset and further classified them into one of the six new classes. Then, an automated process was used to expand this dataset: the process is named Dynamic Query Expansion (DQE)\cite{Dataset1}. This method operates by querying for and labeling, new tweets. Simply put, the top keywords from each class, barring those that ranked highest throughout the entire dataset, were used with the Python library GetOldTweets3 to retrieve and label additional tweets. The resulting dataset contained 69,767 entries. They sampled 8,000 entries for each class from this dataset, resulting in 48,000 data points in total, and thus 40,000 Tweets are classified as Cyberbullying, and 8,000 Tweets as  Not Cyberbullying.\\
\indent Unfortunately, the frequency of data points obtained via DQE was not documented by the researchers, so it is estimated here. Let the number of data points sampled from the original aggregate dataset, that are a result of DQE, be called $S$. This is a discrete random variable described by a hypergeometric distribution. The expected value of this random variable, for any label $l$ is: 
$$E_l[S] = n \dfrac{|S_l|}{N_l}$$
where $n$ is the number of draws, $|S_l|$ is the number of data points from DQE with label $l$, and $N_l$ is the total number of data points with label $l$. Now, as the expected value operation is linear the total expected value is described by:
$$E[S] = \sum\limits_{l \in L}E_l[S].$$
As such, the number of samples resulting from DQE was calculated to be 25,358.20. The significance of this will be discussed in greater detail, in the Methodology section.
It must also be noted that the original dataset contained 205 entries from Dataset 2 (described below). Of course, the sampling performed likely resulted in a smaller number of entries being present in the final dataset. This may have a minor effect on the results of our experiments, but as described later, this supports the conclusions of the experiments.

\subsubsection{Dataset 2}
This dataset was curated with an emphasis on differentiating between offensive language and hate speech \cite{davidson}. They use three labels: Hate Speech, Offensive, and Non-Offensive. The collection process of the tweets was facilitated by the lexicon on Hatebase.org. The timeline of a user who posted a tweet containing a lexical item of the aforementioned lexicon would be collected. The tweets from the timelines were aggregated and this collection was sampled; then tweets were manually labeled. The dataset had 24,802 samples with only 5\% being labeled as Hate Speech. The dataset was curated in 2017 and is available on GitHub. 

\subsubsection{Dataset 3}
The third dataset that we utilize is the “Online Harassment” Dataset, curated in 2017 \cite{Dataset3}. The authors of this dataset initially divided tweets into six classes: The Very Worst, Threats, Hate Speech, Directed Harassment, Potentially Offensive, and Non-Harassing. For simplicity, the tweets were then grouped into two classes such that the first four initial classes were relabeled as Harassing, and tweets of the remaining two classes were labeled as Non-Harassing. Importantly, tweets were collected by using search terms, which, as the authors mention, were related to alt-right / white nationalists. Of the 35,000 tweets collected, over 5,000 were labeled as harassment. The dataset can be requested using the contact info of the authors.

\subsection{Methods} 
This research hypothesizes that the bias introduced from differing definitions of cyberbullying, and from data collection methods, integrally impacts the usefulness of the ML models created and that the performance of these models will drop when applied to other datasets. To test this, cross-validation is used to obtain ML models that will generalize well. Then, these models are tested on unseen datasets. Furthermore, this study formulates the dataset expansion method DQE (See Dataset 1) \cite{Dataset1}, and suggests that algorithmic processes for expanding datasets are potentially harmful when designed or employed improperly.\\

\subsubsection{Theory}
Machine learning models are often described as functions or probabilistic models \cite{Math_for_ML}. Concerning the former, we may understand training machine learning models as the process of discovering, or rather approximating, a function. Thus, if we develop a definition of cyberbullying dependent upon the text within a particular document (e.g. a tweet), we may then utilize the process of ML to approximate a function of document to label. Now, as the methods explored in this study relate to supervised learning, there is the requirement of a labeling process. The labeling process for all datasets in this study, barring a part of Dataset 1, involved human labelers. Thus, if we let X be the set of all tweets and Y, be the set of labels. Then we have two functions to consider:
$$M:X \rightarrow Y, \hspace{.15cm} H:X \rightarrow Y$$
where M is a machine learning model and H is representative of human labelers. Now, the training process can be seen more clearly as an attempt to approximate H with M. Furthermore, a dataset $D$ can be succinctly represented. Let $x \in X_D \subset X$ and $y \in Y_D \subset Y$ then:
$$D = \{(x,y) \hspace{.1cm} | \hspace{.1cm} H(x)=y \}$$
\indent Now, datasets such as the one described above are typically created by obtaining tweets, and then manually labeling them (i.e. applying H to a subset of X). Alternatively, methods such as DQE are used to automate a portion of the labeling process. As discussed in the Datasets section, this method uses labeled tweets to query for additional tweets, and then assigns the label corresponding to the tweets that generated the query, to the new tweet. Grouping dataset entries by their label, we consider the sets of labeled tweets $ D_{All} = \{D_i \hspace{.1cm}| \hspace{.1cm} i \in Y \}$, where $D_i = \{(x,i) \hspace{.1cm}| \hspace{.1cm} H(x)=i \}$ and we formulate Dynamic Query Expansion as:
$$DQE: D_{All} \rightarrow (X,Y), \text{ defined as } DQE(D_y)=(\hat x, y)$$ 
where $\hat x \notin X_D$, is a tweet from an external source and $i = y \in Y_D$ is fixed. Importantly, the label of the resulting data point is the same as the fixed label $y$. Note that this is an oversimplification of the DQE implemented in \cite{Dataset1}, but the relevant consequences still apply. It is now clear that the DQE function simultaneously obtains new tweets and labels them. The method, while removing the burden on human labelers, has an apparent downside and may result in a model approximating the DQE function, rather than the human labelers. Of course, one may argue that DQE is mimicking $H$, and that the model should be approximating it, but this would be an oversight, as if this were the case one might as well use DQE as the predictor function (though it would require modifications), rather than spend time training and maintaining a machine learning model.\\
\indent At this point, it should be clear that this method greatly contributes to the aforementioned issues of biased data collection and bias toward particular terms in a dataset. Importantly, the arguments made in this section apply to any algorithmic process that uses current data points to discover or augment or synthesize data points, not just DQE. Despite these criticisms, DQE may find viable use as a supplemental labeling method in replacement of class balancing techniques such as oversampling, as suggested by the authors \cite{Dataset1}, but in cases like Dataset 1, in which 25,358 samples are estimated to be from DQE, the dataset is likely overtly tainted. If DQE is applied, it is recommended to consider tools such as CleanLab to mitigate its negative effects.\\
\indent Another notable aspect of cyberbullying detection, and NLP in general, is the differing definitions and labeling schemes, used in the labeling process. This can cause problems when attempting to extend the usage of a model trained on one dataset, to another dataset with a different definition of cyberbullying. Rigorously stated, a machine learning model may struggle when it is trained to approximate $H_1$, and then tested on a dataset created via $H_2$. This issue is intuitive but difficult to formulate in-depth.\\
\indent The previous formulations inform the experiments performed in this study and suggest that model performance will likely drop when tested across datasets. As such, we are challenged with creating as many good models as possible, where "good" means that the models will generalize to other datasets. We implement multiple methods to promote the generalizability of the models without letting our prior knowledge of the datasets influence our decisions.

\subsubsection{Preprocessing}
In the interest of time, a few preprocessing steps were applied to all datasets. Links, mentions, retweet symbols, and HTML entities were removed. Additionally, as the datasets are comprised of, almost entirely, English words, tweets with a substantial number of non-English words were removed. Furthermore, as the labels of each dataset differ in name, we converted labels to allow compatibility. The particular labels used by each dataset are converted to one of the two classes: Cyberbullying and Not Cyberbullying. The following conversions are made: all classes except Not Cyberbullying from Dataset 1 are converted to Cyberbullying, the Hate Speech class of Dataset 2 is converted to Cyberbullying, and the Non-Offensive and Offensive classes are converted to Not Cyberbullying, and the Harassing class of Dataset 3 is converted to Cyberbullying and the Non-Harassing class is converted to Not Cyberbullying. The justification of these conversions follows from the given labeling schemes and definitions used to curate these datasets (see the respective papers), though we are aware that this is not the only feasible way to convert the labels.

\subsubsection{Cross Validation}
To perform our experiments, a consistent score is needed to determine the efficacy of the models and ideally inform us which parameters (i.e. models) are likely to generalize well. Acknowledging criticisms of the F1-Score, for not considering the true negative count \cite{F1_criticisms}, these tests utilize the Macro F1 Score as a primary score. The Macro F1 score considers both classes and the reader should note that the true positives of one class are then the true negatives of the other, and as such, the true negatives of both classes are accounted for. Furthermore, as the F1 Score is the harmonic mean of precision and recall, and the Macro F1 Score is the average of this score across all classes, the score is less impacted by class imbalance.\\
\indent Stratified K-Fold cross-validation is repeatedly performed, with different hyperparameters, on each dataset using six splits. To prevent overfitting, a small portion (10\%) of the dataset is set aside to perform early stopping, and standard class weighting is applied (i.e. the positive class is weighted by the ratio of negative samples to positive samples). Additionally, a wide range of parameters, including many that contribute to the prevention of overfitting, are used. The sampling space includes different vectorizers: the CountVectorizer and TfidfVectorizer, as well different different learning rates, max tree depth, etc., for both the CatBoost and XGBoost classifiers (see the code for the full parameter space). The mean of the F1 Macro Score across all six splits is the final score, and the standard deviation of this score is ensured to be low (less than 0.01 for most top results). The models that achieved the highest scores are used in the cross-dataset evaluation. Regrettably, non-stratified cross-validation is not performed.

\subsubsection{Cross-Dataset Evaluation}
The significance of cross-dataset evaluation is not unique to cyberbullying detection \cite{importance_of_cd_testing}, and these experiments attempt to utilize this thorough form of testing. To do so, the top ten models (for both the CatBoost Classifier and the XGB Classifier) that had the highest mean F1 Macro Score during cross-validation, for each dataset, were tested on unseen data (the other two datasets). Thus making the control of these experiments the mean F1 Macro Score achieved during cross-validation. As in the cross-validation, a portion of the dataset is used for early stopping of these models. Each model is then trained on the dataset of interest and then tested on a different dataset. Thus, six experiments in total, two for each dataset, are performed and the F1 Macro Score is recorded, and then compared to the mean F1 Macro score from the respective cross-validation result. The F1 Weighted Score is also recorded and compared with cross-validation results.

\begin{figure}
    \centering
    \includegraphics[width=.8\linewidth]{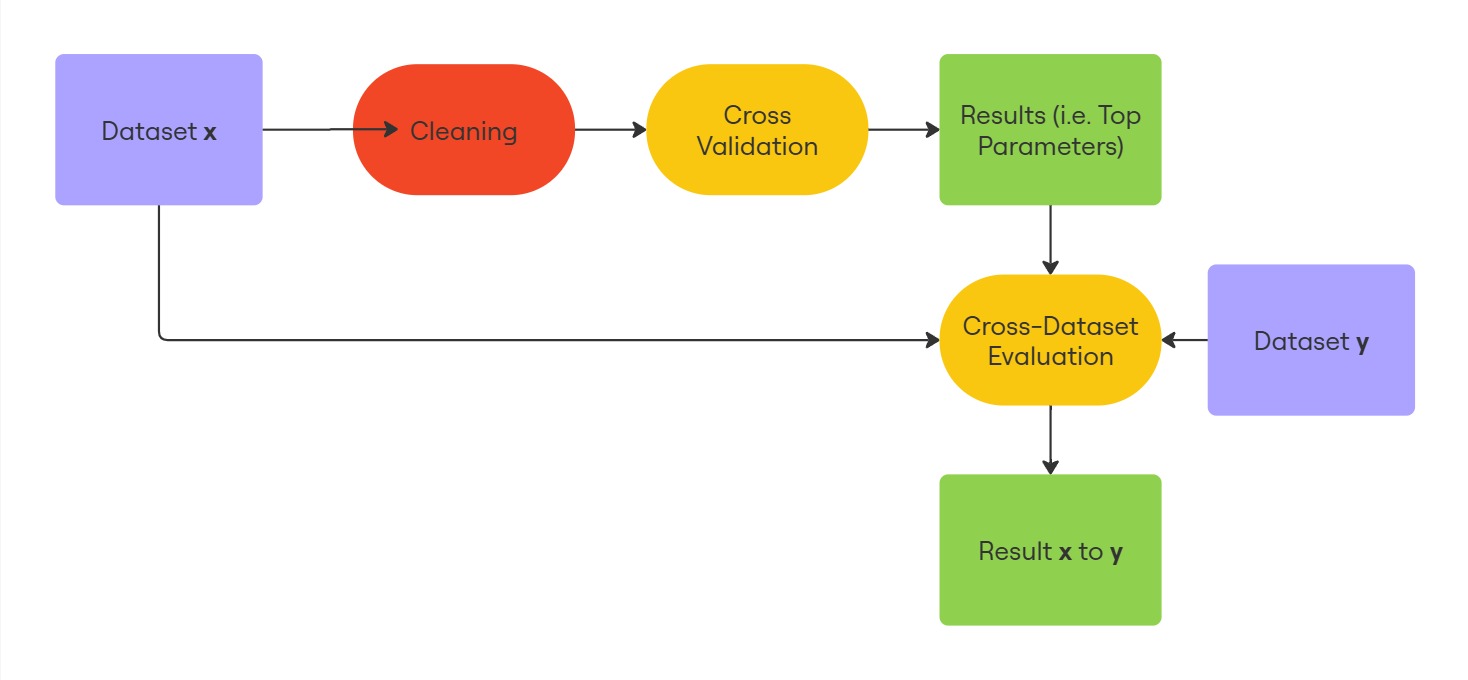}
    \caption{The Experiment Methodology. The process is performed for each Dataset \textbf{x} and for each of the remaining datasets \textbf{y} such that six total experiments are performed. Note that the Cleaning, Cross Validation. and Cross-Dataset Evaluation only vary with regard to the inputs depicted.}
    \label{fig:exp_process}
\end{figure}

\section{Results}
\label{sec:Results}

\subsection{Cross Validation Results}
\begin{figure}[H]
\centering
\begin{subfigure}{0.5\textwidth}
\centering
    \includegraphics[scale=0.43]{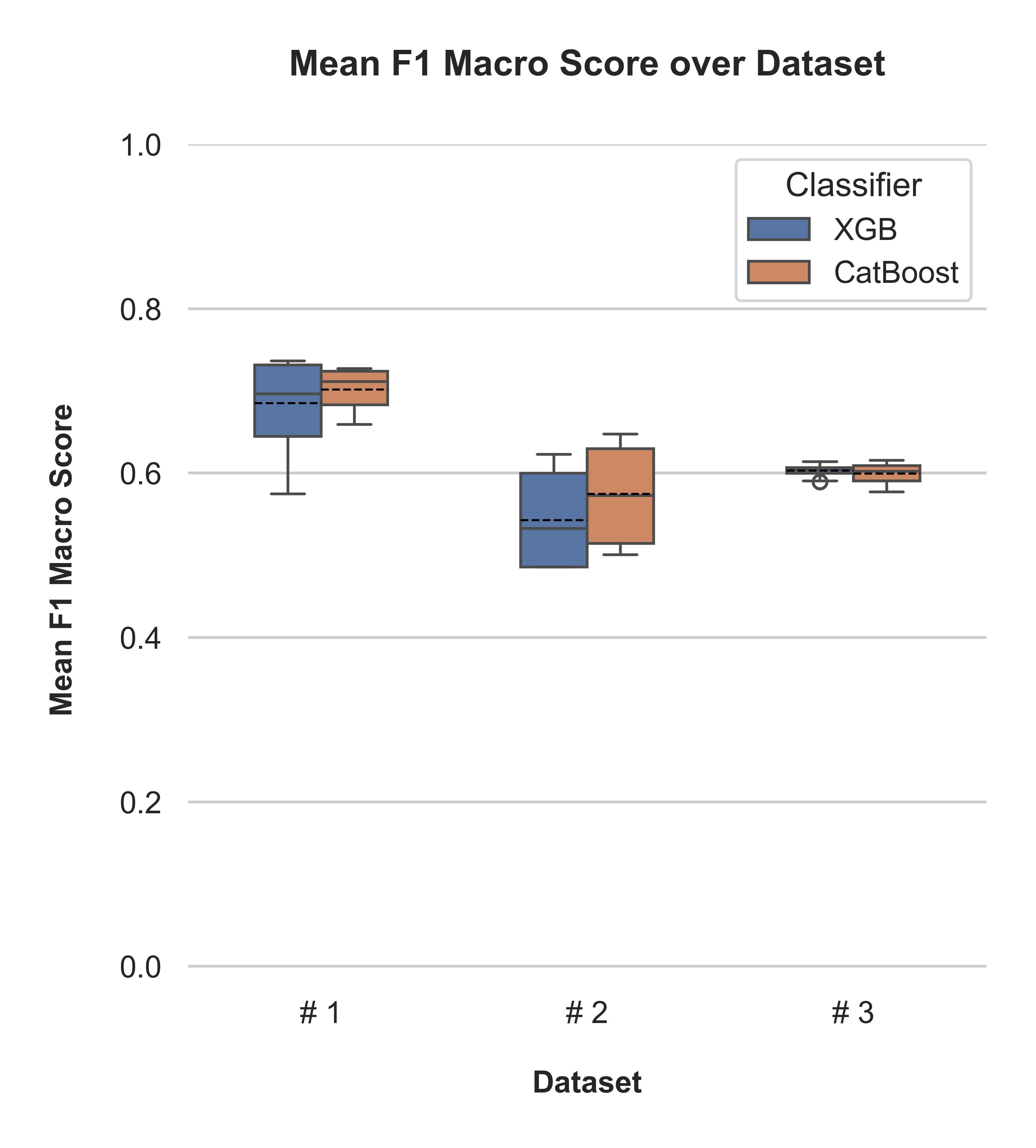}
    \label{fig:m_f1_macro}
\end{subfigure}%
\begin{subfigure}{0.5\textwidth}
\centering
    \includegraphics[scale=0.43]{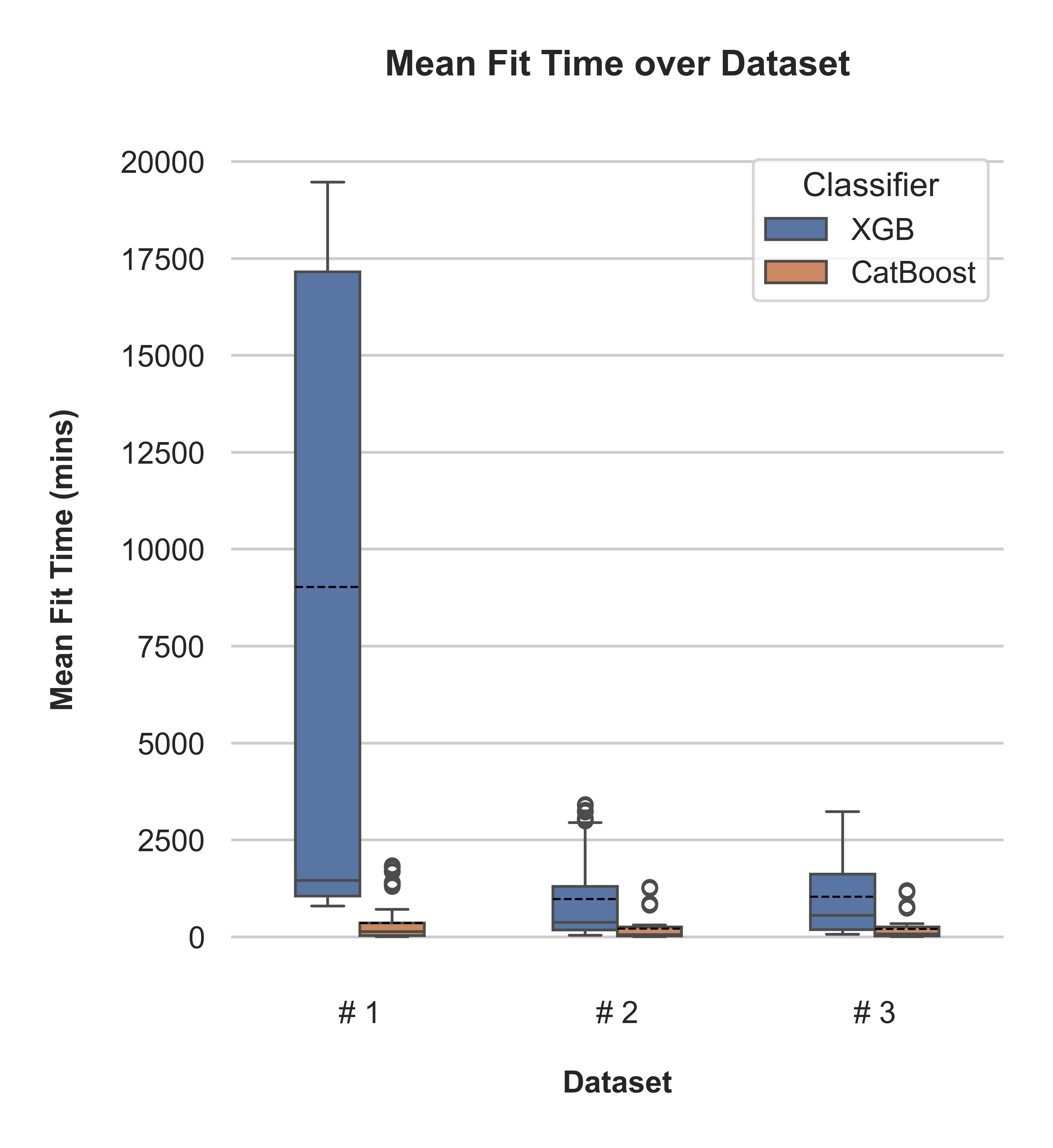}
    \label{fig:m_fit_time}
\end{subfigure}
\caption{Cross Validation Results}
\label{fig:cv_results}
\end{figure}

Figure~\ref{fig:cv_results} displays the mean F1 Macro Score and the mean Fit Time, for each dataset and classifier type. Many of the models did not perform particularly well during cross-validation, suggesting that they will not generalize well. However, many of the models for dataset 1 did achieve a moderate score of 0.7 or above, suggesting that these models may have the potential to generalize well. Also, the fit time of XGBoost models during cross-validation of dataset 1 was very costly, and while diagnosis of this issue is not in the scope of this research, it is a topic of interest for future research.

\subsection{Cross-Dataset Evaluation Results}
\begin{figure}[H]
\centering
\begin{subfigure}{0.5\textwidth}
\centering
    \includegraphics[scale=0.43]{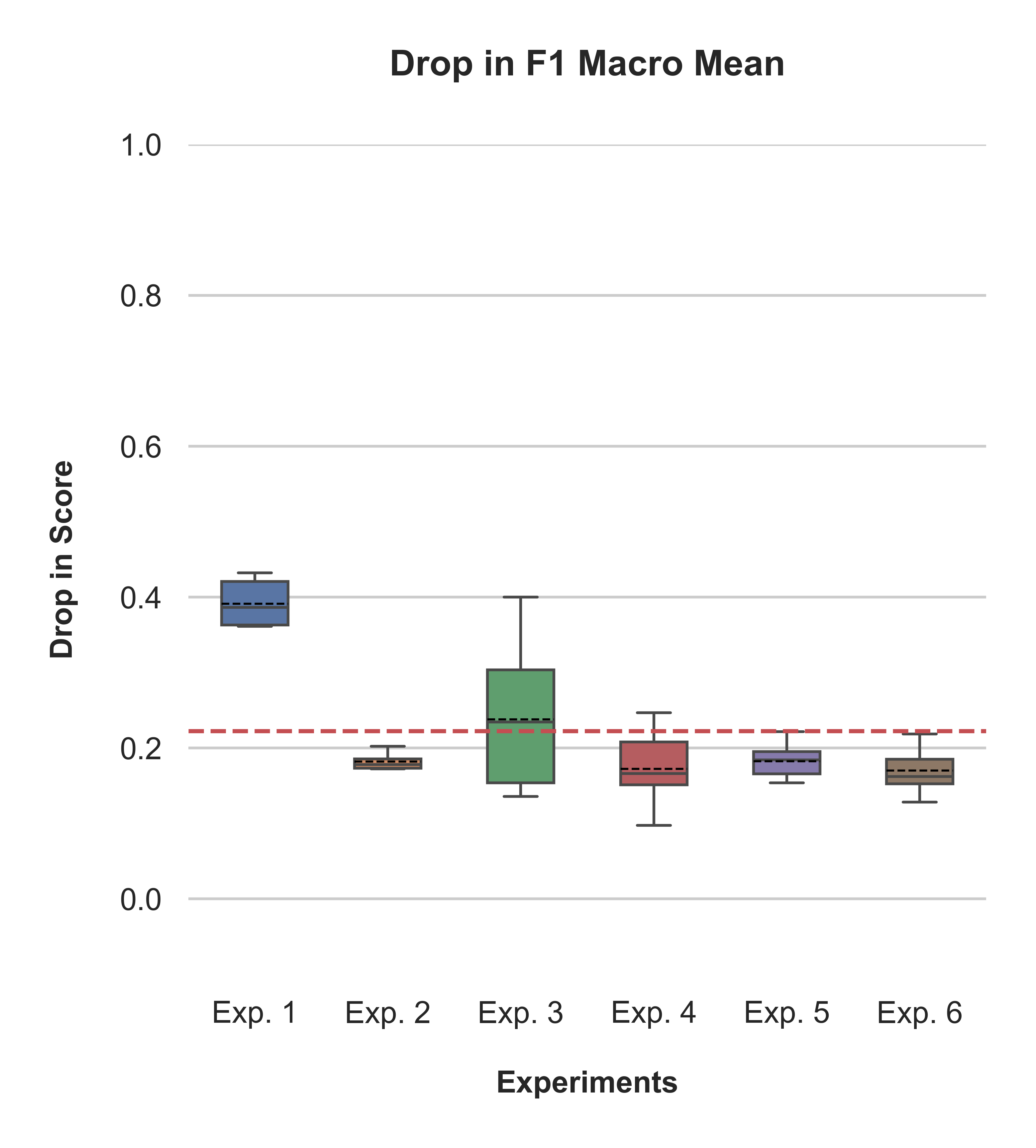}
    \label{fig:f1_m_drop}
\end{subfigure}%
\begin{subfigure}{0.5\textwidth}
\centering
    \includegraphics[scale=0.43]{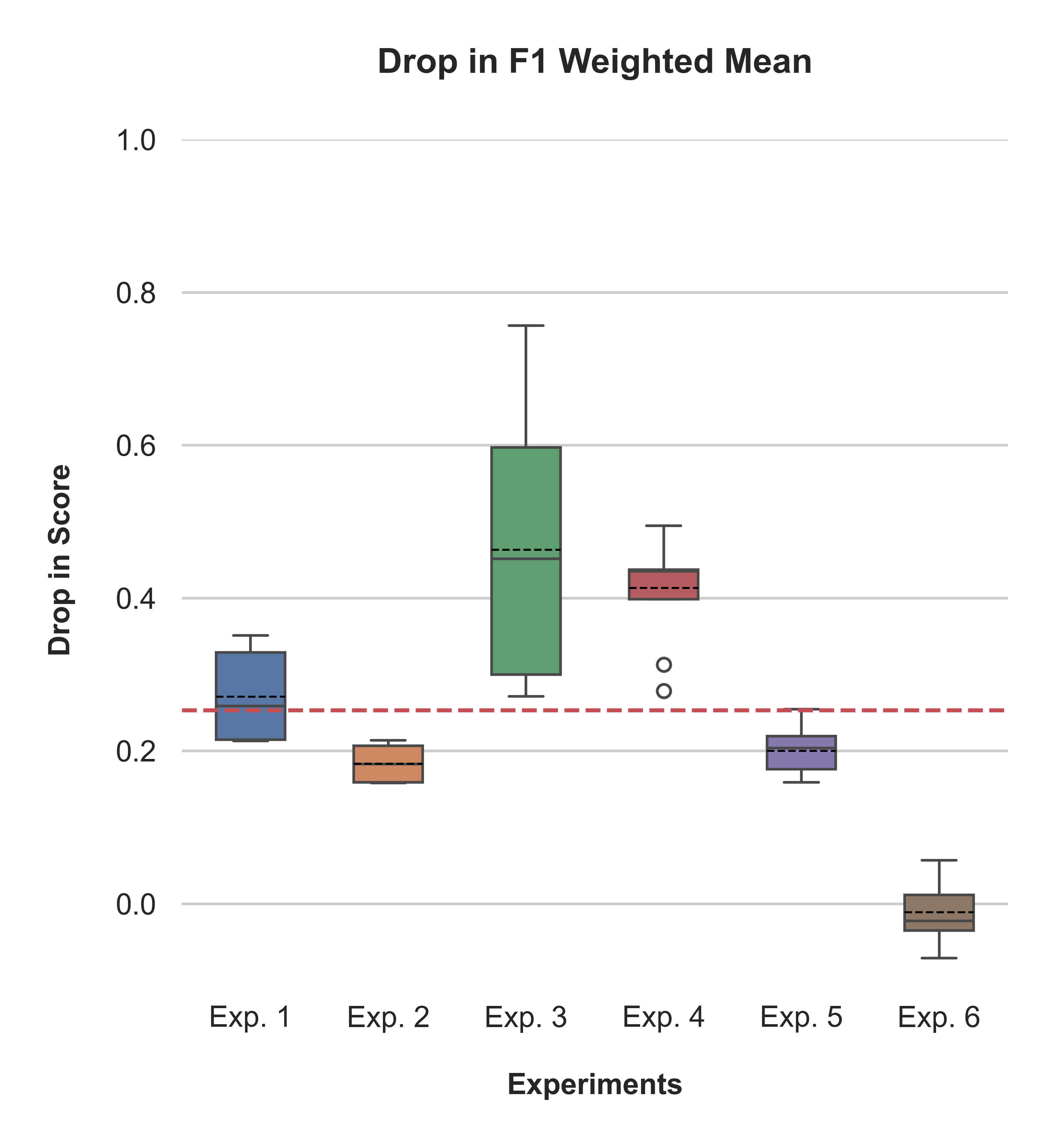}
    \label{fig:f1_w_drop}
\end{subfigure}
\caption{Cross-Dataset Evaluation Results. The red dashed line indicates the average across all experiments.}
\label{fig:cde_results}
\end{figure}

Figure~\ref{fig:cde_results} displays the drop, from the cross-validation to the cross-dataset evaluation, in the Macro F1 Score and the F1 Weighted Score, respectively. The results are split into six experiments (Table~\ref{tab1}). Also, each experiment only uses the top 10 models that scored highest during cross-validation, for the respective dataset. Despite performing moderately during cross-validation, models trained on dataset 1 had a performance drop comparable to, or worse than, models trained on the other datasets. The red-dashed line in the figure depicts the average drop across all experiments and the black-dashed lines on each bar graph depict the average drop for that particular experiment (Figure~\ref{fig:cde_results}).

\begin{table}[h] 
\centering
\caption{Numerical representations of the cross-dataset experiment results.\label{tab1}}
\newcolumntype{C}{>{\centering\arraybackslash}X}
%\scalebox{0.8}{
\begin{tabularx}{\textwidth}{CCCCC}
\toprule
\textbf{Experiment Number}	& \textbf{Training Set}	& \textbf{Testing Set} & \textbf{Avg. Drop in F1 Macro} & \textbf{Avg. Drop in F1 Weighted}\\
\midrule
Exp. 1  &   Dataset 1   &   Dataset 2   &   0.3912   &   0.2710   \\
Exp. 2  &   Dataset 1   &   Dataset 3   &   0.1819   &   0.1835   \\
Exp. 3  &   Dataset 2   &   Dataset 1   &   0.2379   &   0.4634   \\
Exp. 4  &   Dataset 2   &   Dataset 3   &   0.1724   &   0.4133   \\
Exp. 5  &   Dataset 3   &   Dataset 1   &   0.1822   &   0.2000   \\
Exp. 6  &   Dataset 3   &   Dataset 2   &   0.1699   &  -0.0108   \\
\bottomrule
\end{tabularx}
\end{table}

%%%%%%%%%%%%%%%%%%%%%%%%%%%%%%%%%%%%%%%%%%
\section{Discussion}

Despite the usage of overfitting prevention methods such as early stopping, class weighting, and setting parameters specific to the model and vectorizer types, it is easily observed that the models perform poorly in the cross-dataset evaluation compared to their performance during cross-validation. As indicated by the dashed red line, the average drop in the Macro Mean F1 score was 0.222 and the average drop in the Weighted Mean F1 score was 0.253, across all experiments (Figure~\ref{fig:cde_results}). This evidence clearly supports our hypothesis, but further analysis is required to determine the cause of these results.\\

\indent One method used to determine a potential cause of this drop in performance was correlation tests. As such, to gauge the dependency of the drop in F1 Score on out-of-vocabulary terms, the Kendall-Tau and Spearman correlation coefficients were calculated (Tables~\ref{tab2} and~\ref{tab3}). The correlation tests done are between the ratio of out-of-vocabulary (OOV) terms to the total terms in a dataset, $R$, and the drop in the F1 Macro Mean and F1 Weighted Mean, respectively. The results suggest a negative correlation between $R$ and the drop in the F1 Macro Mean and no correlation with the drop in the F1 Weighted Mean. Importantly, both results are supportive of the conclusion that there is no positive correlation between $R$ and the drop in F1 Score. This is counterintuitive, as we know that when $R=1$, models would not be able to make predictions and there would be a 100\% drop in prediction scores. In our case though, the test results suggest that $R$ was not high enough to have a consistent impact on the drop in scores, and as such we conclude that the drop in scores, for our experiments, must be due to other factors.

\begin{table}[H] 
\caption{Correlation Test Results between $R$ and the Drop in the F1 Macro Score\label{tab2}}
\newcolumntype{C}{>{\centering\arraybackslash}X}
\begin{tabularx}{\textwidth}{CCC}
\toprule
\textbf{Correlation Type}	& \textbf{Correlation Coefficient}	& \textbf{p-value}\\
\midrule
Kendall-Tau		& -0.1587  	& 0.0116\\
Spearman		& -0.2190	& 0.0163\\
\bottomrule
\end{tabularx}
\end{table}

\begin{table}[H] 
\caption{Correlation Test Results between $R$ and the Drop in the F1 Weighted Score\label{tab3}}
\newcolumntype{C}{>{\centering\arraybackslash}X}
\begin{tabularx}{\textwidth}{CCC}
\toprule
\textbf{Correlation Type}	& \textbf{Correlation Coefficient}	& \textbf{p-value}\\
\midrule
Kendall-Tau		& 0.0071	& 0.9096\\
Spearman		& -0.0021	& 0.9823\\
\bottomrule
\end{tabularx}
\end{table}

\section{Conclusions} 
Despite performing well, or moderately, during cross-validation, we observed a sufficient decrease in model performance during the cross-dataset evaluation. We do not explicitly provide evidence that differing labeling schemes or definitions affect the overall performance drop, but we have thoroughly argued against the many biased methodologies employed in cyberbullying detection. The results of experiments on biased data, while worth studying, are misleading with regard to the curated models' abilities. This is not to say that these models are unusable, but rather it is strong evidence of a need to understand the data collection, definitions, and labeling scheme used during training and testing. To apply these models without an understanding of the data will likely result in unexpected results and poor performance, as we have shown. Considering that cyberbullying detection will likely be used as a means to censor communication, potentially without just cause, it is in the best interest of those creating machine learning models to understand when a model, or a dataset, is truly applicable. As for future work, we will continue on the path of not only mitigating bias but understanding it.  

% \subsection{Citations}
% Citations use \verb+natbib+. The documentation may be found at
% \begin{center}
% 	\url{http://mirrors.ctan.org/macros/latex/contrib/natbib/natnotes.pdf}
% \end{center}

% Here is an example usage of the two main commands (\verb+citet+ and \verb+citep+): Some people thought a thing \citep{kour2014real, hadash2018estimate} but other people thought something else \citep{kour2014fast}. Many people have speculated that if we knew exactly why \citet{kour2014fast} thought this\dots

% \subsection{Figures}
% \lipsum[10]
% See Figure \ref{fig:fig1}. Here is how you add footnotes. \footnote{Sample of the first footnote.}
% \lipsum[11]

% \begin{figure}
% 	\centering
% 	\fbox{\rule[-.5cm]{4cm}{4cm} \rule[-.5cm]{4cm}{0cm}}
% 	\caption{Sample figure caption.}
% 	\label{fig:fig1}
% \end{figure}

% \subsection{Tables}
% See awesome Table~\ref{tab:table}.

% The documentation for \verb+booktabs+ (`Publication quality tables in LaTeX') is available from:
% \begin{center}
% 	\url{https://www.ctan.org/pkg/booktabs}
% \end{center}

% \begin{table}
% 	\caption{Sample table title}
% 	\centering
% 	\begin{tabular}{lll}
% 		\toprule
% 		\multicolumn{2}{c}{Part}                   \\
% 		\cmidrule(r){1-2}
% 		Name     & Description     & Size ($\mu$m) \\
% 		\midrule
% 		Dendrite & Input terminal  & $\sim$100     \\
% 		Axon     & Output terminal & $\sim$10      \\
% 		Soma     & Cell body       & up to $10^6$  \\
% 		\bottomrule
% 	\end{tabular}
% 	\label{tab:table}
% \end{table}

% \subsection{Lists}
% \begin{itemize}
% 	\item Lorem ipsum dolor sit amet
% 	\item consectetur adipiscing elit.
% 	\item Aliquam dignissim blandit est, in dictum tortor gravida eget. In ac rutrum magna.
% \end{itemize}

\newpage
\bibliographystyle{IEEEtran}
\bibliography{references}  %%% Uncomment this line and comment out the ``thebibliography'' section below to use the external .bib file (using bibtex) .





% \end{thebibliography}

\end{document}